\documentclass[10pt,twocolumn,letterpaper]{article}
\usepackage[accsupp]{axessibility}

\usepackage{cvpr}

\usepackage{algorithm}
\usepackage{amsmath}

\usepackage{algpseudocode} 
\usepackage{multirow}
\usepackage{makecell}
\usepackage{multirow}
\usepackage{booktabs}
\usepackage{xcolor}

\definecolor{cvprblue}{rgb}{0.21,0.49,0.74}
\usepackage[pagebackref,breaklinks,colorlinks,allcolors=cvprblue]{hyperref}

\title{Unsupervised Multi-agent and Single-agent Perception from Cooperative Views}

\author{
Haochen Yang\textsuperscript{1} \quad 
Baolu Li\textsuperscript{1} \quad 
Lei Li\textsuperscript{1} \quad 
Delin Ren\textsuperscript{1} \quad 
Jiacheng Guo\textsuperscript{1} \quad 
Minghai Qin\textsuperscript{1} \quad \\
Tianyun Zhang\textsuperscript{1}$^{*}$ \quad 
Hongkai Yu\textsuperscript{1}$^{*}$ \\
\textsuperscript{1}Cleveland State University
}

\begin{document}
\maketitle

{\renewcommand{\thefootnote}{}\footnotetext{
$^{*}$Corresponding authors: Tianyun Zhang (t.zhang85@csuohio.edu), Hongkai Yu (h.yu19@csuohio.edu).
}}

\begin{abstract}

The LiDAR-based multi-agent and single-agent perception has shown promising performance in environmental understanding for robots and automated vehicles.  However, there is no existing method that simultaneously solves both multi-agent and single-agent perception in an unsupervised way. By sharing sensor data between multiple agents via communication, this paper discovers two key insights: 1) Improved point cloud density after the data sharing from 
cooperative views could benefit unsupervised object classification, 2) Cooperative view of multiple agents can be used as unsupervised guidance for the 3D object detection in the single view. Based on these two discovered insights, we propose an \textbf{U}nsupervised \textbf{M}ulti-agent and \textbf{S}ingle-agent (\textbf{UMS}) perception framework that leverages multi-agent cooperation \textbf{without human annotations} to simultaneously solve multi-agent and single-agent perception. UMS combines a learning-based Proposal Purifying Filter to better classify the candidate proposals after multi-agent point cloud density cooperation, followed by a Progressive Proposal Stabilizing module to yield reliable pseudo labels by the easy-to-hard curriculum learning. Furthermore, we design a Cross-View Consensus Learning to use multi-agent cooperative view to guide detection in single-agent view. Experimental results on two  public datasets  V2V4Real and OPV2V  show that our UMS method achieved significantly higher 3D detection  performance than the state-of-the-art methods on both multi-agent and single-agent perception tasks in an unsupervised setting.

\end{abstract}
    
\section{Introduction}
\label{sec:introduction}

\begin{figure}[t]
    \centering
    \includegraphics[width=0.478\textwidth]{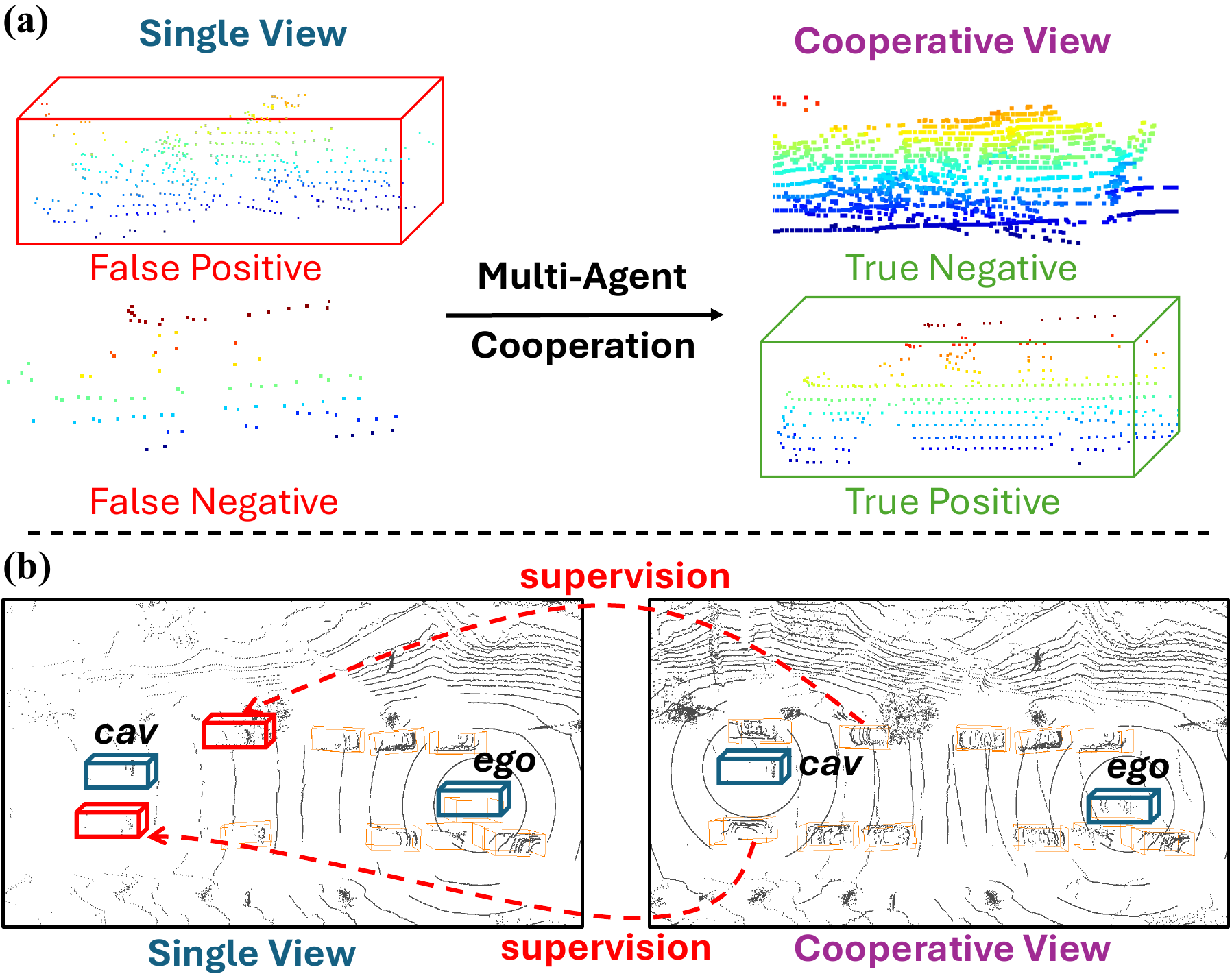}
    \caption{\textbf{Illustration of Benefits from Cooperative Views.} (a) Point Cloud Density Benefit, (b) Cross-View Consensus Benefit. We use Vehicle-to-Vehicle (V2V) cooperative perception~\cite{xu2022opv2v} in Connected Autonomous Vehicles (CAV) as an example here. 
    }
    \label{fig:intro0}
\end{figure}

Multi-agent cooperative perception~\cite{xu2022opv2v} has displayed more advanced 3D object detection than single-agent perception, because the multi-agent system greatly enlarges the perception range by aggregating the shared point cloud data of multiple agents. Either multi-agent or single-agent perception highly relies on the intensive supervised learning from large-scale human‑annotated 3D bounding boxes of the objects, which are always not available in many real-world applications. Therefore, is it possible to jointly solve multi-agent and single-agent perception only by the communication between agents without human annotations?

A naive answer is to use prior knowledge of communicated agents to automatically generate 3D bounding boxes of communicated agents, which can be used to train a 3D object detector. Although point cloud based physical rules~\cite{xia2025learning} can slightly refine detection errors, the naive method will still produce a large number of false positives and false negatives in detection. To solve this problem, as shown in Fig.~\ref{fig:intro0}, we observe the following key insights to benefit unsupervised learning from cooperative views.

\begin{itemize}
    \item \textbf{Point Cloud Density Benefit}:  As shown in Fig.~\ref{fig:intro0}(a), improved point cloud density by multi-agent cooperation (data sharing) from cooperative views makes unsupervised 3D object classification easier, e.g., the binary classification of Vehicle and Non-Vehicle. 
    
    \item \textbf{Cross-View Consensus Benefit}: As shown in Fig.~\ref{fig:intro0}(b), complementary consensus cues from the multi-agent cooperative view provide unsupervised guidance to 3D object detection in single-agent view.
\end{itemize}

To the best of our knowledge, there is no existing method that simultaneously solves both multi-agent and single-agent perception in an unsupervised way. Based on the above two discovered benefits, we propose a novel  \textbf{U}nsupervised \textbf{M}ulti-agent and \textbf{S}ingle-agent (\textbf{UMS}) perception framework that leverages multi-agent cooperation without human annotations to simultaneously solve multi-agent and single-agent perception. Inspired by the Point Cloud Density Benefit in Fig.~\ref{fig:intro0}(a), we design a new hierarchical point cloud feature learning based Proposal Purifying Filter (PPF) to better classify the candidate proposals after multi-agent point cloud density cooperation. In addition, to yield reliable pseudo labels, we design a new Progressive Proposal Stabilizing (PPS) module to stabilize the object proposals by the idea of easy-to-hard curriculum learning~\cite{bengio2009curriculum}. Furthermore, inspired by the Cross-View Consensus Benefit in Fig.~\ref{fig:intro0}(b), we develop a new Cross-View Consensus Learning (CCL) involving Multi-View Geometric Consensus and BEV (Bird's-Eye View) Semantic Alignment to use multi-agent cooperative view to guide the object detection in single-agent view. Finally, our UMS method is tested on two public datasets V2V4Real~\cite{xu2023v2v4real} (real world) and OPV2V~\cite{xu2022opv2v} (simulated) for Vehicle-to-Vehicle (V2V) multi-agent cooperative perception and single-agent perception, which performs  impressively better 3D object detection than other state-of-the-art methods under the  unsupervised settings. The main contributions of this paper are summarized as follows.

\begin{itemize}
    \item  We propose the first method (named as UMS) to simultaneously solve both multi-agent and single-agent perception in an unsupervised way, which works well for an ego agent with or without communication based data sharing with other agents.

    \item  By the enhanced point cloud density from cooperative views, we propose two new modules Proposal Purifying Filter (PPF) and  Progressive Proposal Stabilizing (PPS) to promote object recognition in the multi-agent perception. 

    \item By the complementary consensus cues from cooperative views, we propose a new module Cross-View Consensus Learning (CCL), including Multi-View Geometric Consensus and BEV Semantic Alignment, to improve object detection in the single-agent perception. 
\end{itemize}

\section{Related Work}

\begin{figure*}[htbp]
    \centering
    \includegraphics[width=\textwidth]{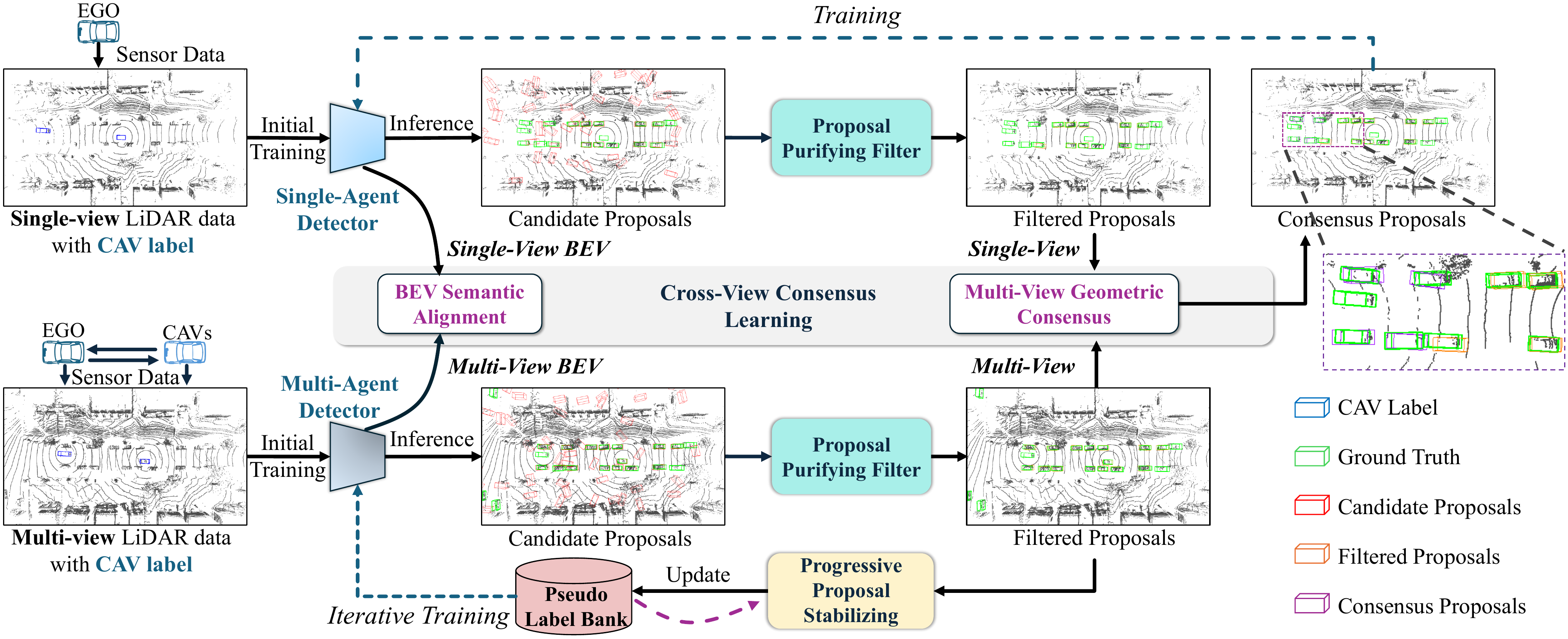}
    \caption{\textbf{UMS Pipeline.} The system jointly trains two 3D object detectors for multi-agent and single-agent  perception. Candidate Proposals are first generated from two initialized weak detectors and then refined by (i) Proposal Purifying Filter (PPF), (ii) Progressive Proposal Stabilizing (PPS), and (iii) Cross-View Consensus Learning (CCL), enabling robust pseudo supervision without human annotations.}
    \label{fig:overall}
\end{figure*}

\subsection{Multi-agent Cooperative Perception}

Cooperative perception has been extensively studied to address the limitations of single-agent perception. By enabling information exchange between different agents, cooperative systems increase perception coverage and improve scene understanding in complex traffic scenarios. Existing methods have explored various fusion strategies~\cite{li2021learning,xiang2022v2xp,hu2023collaboration,xu2022opv2v,li2024s2r,li2025v2x,xu2023v2v4real}, especially intermediate fusion, due to its favorable trade-off between perception performance and communication~\cite{ hu2024communication,lei2022latency,wei2023asynchrony}. However, most previous studies focus primarily on improving cooperative perception itself, while overlooking practical deployment scenarios where the penetration of V2V-enabled vehicles remains limited, leaving many vehicles to rely on single-agent perception~\cite{albattah2022overview, huang2025vehicle}. Moreover, existing approaches heavily depend on large-scale annotated datasets, which are costly to obtain and difficult to scale to new environments~\cite{xu2023v2v4real}. Therefore, how to leverage limited cooperative agents to improve both multi-agent and single-agent 3D perception without manual annotations remains underexplored.

\subsection{Unsupervised 3D Object Detection}
 
Unsupervised 3D object detection aims to train detectors without manual annotations. Existing methods mostly work in single-agent settings, generating pseudo labels from motion cues or time consistency~\cite{you2022learning,najibi2022motion,tian2021unsupervised}. Recent work explores heuristic or clustering-based strategies. OYSTER~\cite{zhang2023towards} groups points by clustering but struggles to separate sparse objects from the structured background. CPD~\cite{wu2024commonsense} introduces prototype templates and commonsense-driven filtering to improve robustness, but it remains sensitive to incomplete observations. DOtA~\cite{xia2025learning} extends to multi-agent settings, but its rule-based filtering degrades noticeably under occlusion scenes. Multimodal methods~\cite{lentsch2024union,wang20224d,fruhwirth2024vision} offer complementary cues but do not resolve the geometry missing problem caused by sparse LiDAR, which limits detection accuracy. Unlike the above methods, this paper exploits cooperative LiDAR to get denser point clouds and enforces cross-view consistency as supervision, building a learnable framework that generates more reliable pseudo labels for both multi-agent and single-agent perception.

\section{Method}
\label{sec:method}

\subsection{Overall Framework}

Using V2V cooperation as study example, $V$ connected vehicles provide shared LiDAR point clouds $\{X^v\}_{v=1}^V$ together with GPS poses~\cite{wang2020v2vnet}. All point clouds are transformed into the ego-view coordinate system of the single-agent vehicle $X^{e}$ using Homogeneous Transformation Matrix $\{T_{v\rightarrow e}\}$. We maintain two detectors in training: a single-agent detector $D_{\mathrm{e}}$ (without data sharing) and a multi-agent detector $D_{\mathrm{m}}$ (with data sharing). They are optimized through iterative pseudo-supervision.

Without human annotations, the  positional priors (location, pose) of communicated vehicles are used to train two weak detectors $D_{\mathrm{m}}, D_{\mathrm{e}}$, providing  initialized pseudo labels as start. The initialized pseudo labels are \textit{3D bounding boxes for objects, which are also called proposals in this paper}. The weak single-agent detector $D_{\mathrm{e}}$ takes $X^{e}$ and outputs candidate proposals $\mathcal{P}_{\mathrm{e}}=\{b^{\mathrm{e}}, c^{\mathrm{e}}\}$. The weak multi-agent detector $D_{\mathrm{m}}$ takes $\{X^v\}_{v=1}^{V}$ and outputs candidate proposals  $\mathcal{P}_{\mathrm{m}}=\{b^{\mathrm{m}}, c^{\mathrm{m}}\}$. Each bounding box is defined as $b$ to  describe a 3D object, and $c$ is its confidence. These proposals are inevitably noisy and incomplete by the initialized weak detectors.

As shown in Fig.~\ref{fig:overall}, our method refines these candidate proposals in three stages. The Proposal Purifying Filter (PPF) removes unreliable boxes and produces filtered results $\widetilde{\mathcal{P}}_{\mathrm{e}}$ and $\widetilde{\mathcal{P}}_{\mathrm{m}}$. The Progressive Proposal Stabilizing (PPS) module fuses $\widetilde{\mathcal{P}}_{\mathrm{m}}$ with historical proposals in a memory bank $\mathcal{B}$ to obtain stabilized results $\widehat{\mathcal{P}}_{\mathrm{m}}$. These results improve stability and recall. The Cross-View Consensus Learning (CCL) module then enforces geometric and semantic consistency between the two detectors using geometric and BEV features. It also generates consensus pseudo labels $\widehat{\mathcal{P}}_{\mathrm{e}}$ to train $D_{\mathrm{e}}$. By these refinements, the initialized weak detectors $D_{\mathrm{m}}, D_{\mathrm{e}}$ can be enhanced.

\subsection{Proposal Purifying Filter}
\label{sec:ppf}

\begin{figure*}[!t]
    \centering
    \includegraphics[width=\textwidth]{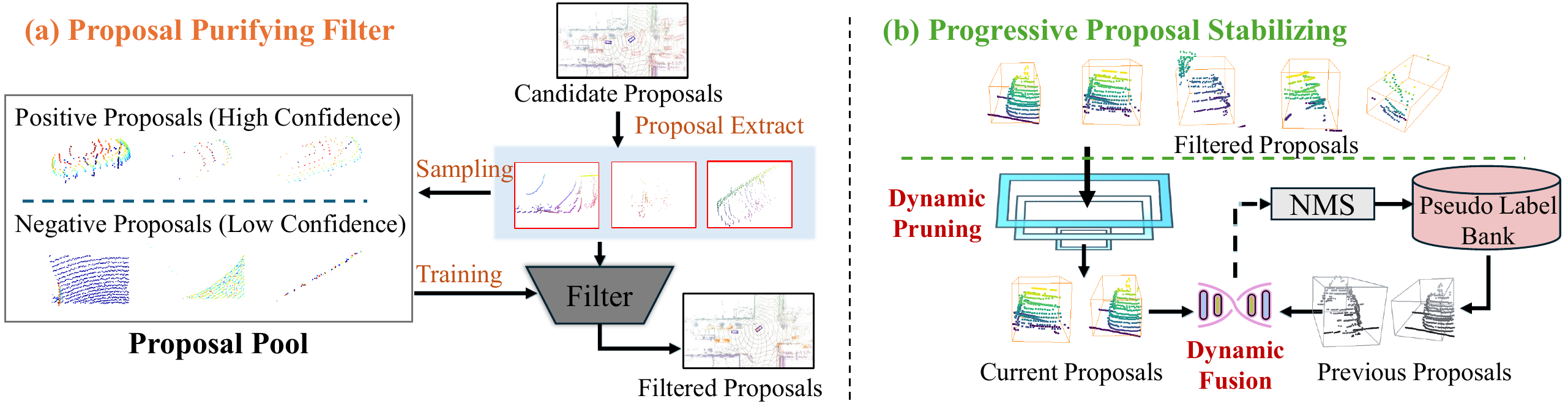}
    \caption{\textbf{Illustration of the proposed modules.}
    (a) \textbf{Proposal Purifying Filter (PPF)} learns an instance-level filter/classifier to remove unreliable proposals.
    (b) \textbf{Progressive Proposal Stabilizing (PPS)} maintains a memory bank and adaptively fuses historical and current pseudo labels for stability.
    }
    \label{fig:method1}
    \vspace{-1em}
\end{figure*}

Multi-agent cooperation provides denser point clouds after data sharing. Existing clustering-based methods rely on hand-crafted heuristics (e.g., DBSCAN~\cite{ester1996density}, OYSTER~\cite{zhang2023towards}, CPD~\cite{wu2024commonsense}) and lack instance-level feature learning. A learnable filter/classifier is desired, but the lack of human annotations raises challenges. To address this challenge, we propose Proposal Purifying Filter (PPF), in Fig.~\ref{fig:method1}(a), to learn hierarchical point cloud features at the instance level to facilitate the self-supervised classification.

\begin{figure}[htbp]
    \centering
\includegraphics[width=0.40\textwidth]{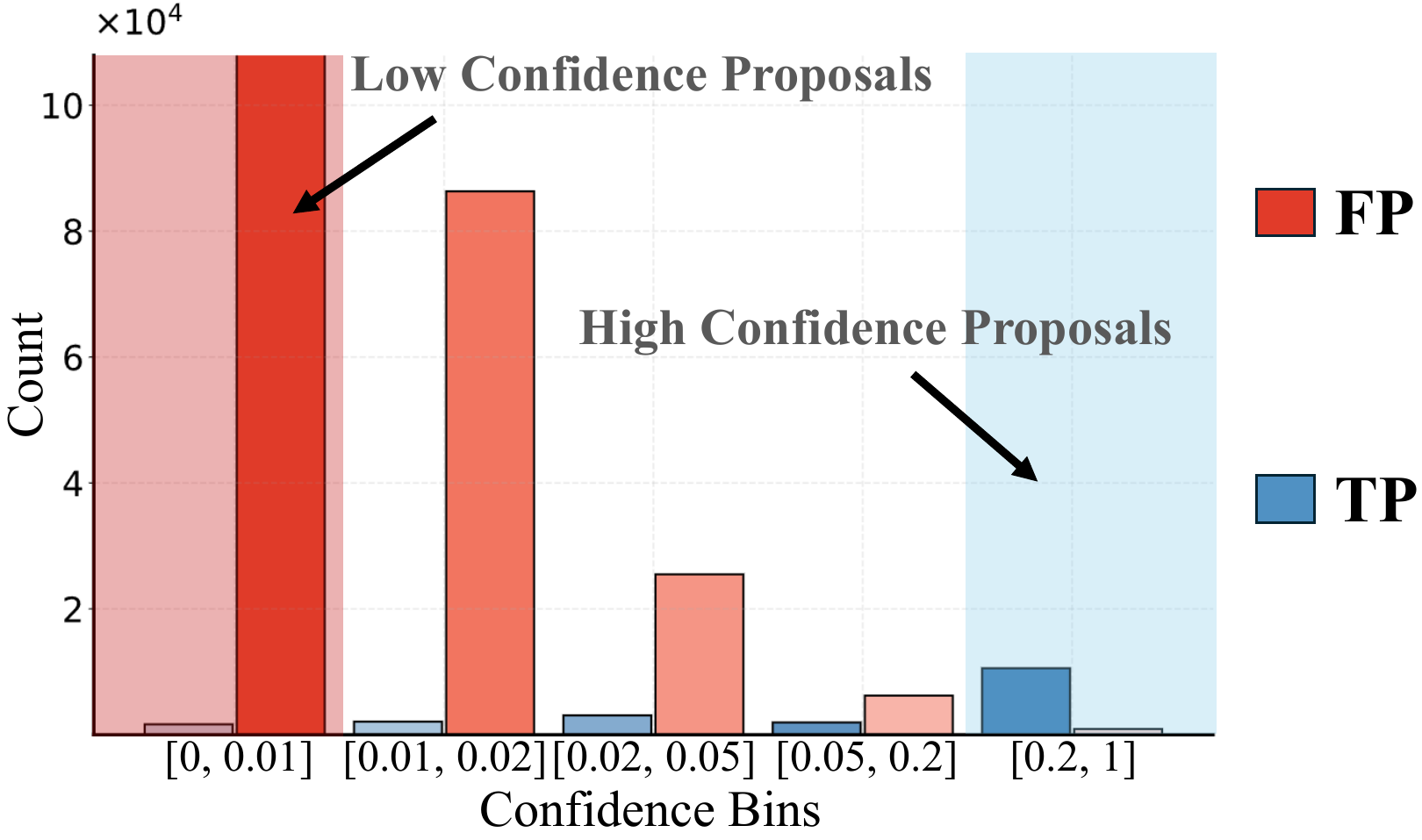}
\caption{\textbf{Confidence–TP/FP statistics with improved point cloud density under multi-agent setting.}
Based on the initialized weak detector 
$D_{\mathrm{m}}$ on the V2V4Real~\cite{xu2023v2v4real} training set, the distribution gap of True Positives (TP) and False Positives (FP) across confidence makes self-supervised classification possible.}
    \label{fig:cgif1}
\end{figure}

\noindent\textbf{Training.}
By the initialized weak detector $D_{\mathrm{m}}$, high-confidence proposals are mostly true positives, while low-confidence ones are mostly false positives, as shown in Fig.~\ref{fig:cgif1}. These two groups provide self-supervision for classification. Among all proposals in $\mathcal{P}_{\mathrm{m}}$, we select a negative set $\mathcal{S}^{-}$ with low-confidence proposals and a positive set $\mathcal{S}^{+}$ with high-confidence proposals. For each proposal $b_i$ in $\mathcal{S}^{-}\cup\mathcal{S}^{+}$, we define $\mathrm{crop}(X, b_i)$ as the operation that extracts all points within the proposal $b_i$ from the full point cloud $X$. We feed $\mathrm{crop}(X, b_i)$ into an instance-level  hierarchical point cloud feature extraction (PointNet++~\cite{qi2017pointnet++}) based classifier $C_{\phi}$ that predicts a classification score:
\begin{equation}
q_i = C_{\phi}\!\left(\mathrm{crop}(X, b_i)\right), 
\label{eq:ppf}
\end{equation}
where $q_i \in [0,1]$. Each sample is assigned a binary target label $y_i\in\{0,1\}$ based on its membership in $\mathcal{S}^{-}$ or $\mathcal{S}^{+}$. The classifier $C_{\phi}$ is trained with Binary Cross Entropy loss: 
\begin{equation}
\mathcal{L}_{\mathrm{ppf}} = -\sum_{i}[y_i\log q_i + (1 - y_i)\log(1 - q_i)].
\label{eq:ppf_loss}
\end{equation}
\noindent\textbf{Testing.}
After training, PPF evaluates each proposal $b_i$ in $\mathcal{P}_{\mathrm{e}}$ and $\mathcal{P}_{\mathrm{m}}$ using Eq.~\eqref{eq:ppf}. A proposal is retained only if $q_i \ge 0.5$. This produces two filtered sets $\widetilde{\mathcal{P}}_{\mathrm{e}}$ and $\widetilde{\mathcal{P}}_{\mathrm{m}}$.

By learning instance-level hierarchical point cloud features in dense point clouds under multi-agent setting, PPF effectively suppresses clutters caused by  sparse point clouds in the single-agent scenario, which is also obviously stronger than hand-crafted heuristics~\cite{ester1996density,zhang2023towards,wu2024commonsense}.

\subsection{Progressive Proposal Stabilizing}
\label{sec:pps}

Although PPF removes unreliable proposals, the remaining ones may still fluctuate due to intermittent visibility, viewpoint change, and sparse points. Thus, we propose an iterative way to perform progressive proposal stabilizing. Proposals persisting across iterations are more likely to be real objects, whereas noisy ones are short-lived. Inspired by easy-to-hard curriculum learning~\cite{bengio2009curriculum}, which gradually involves hard examples during training, we introduce Progressive Proposal Stabilizing (PPS) as shown in Fig.~\ref{fig:method1} (b). PPS employs two dynamic mechanisms: (i) a \emph{Dynamic Pruning} that starts with low confidence to admit more pseudo labels early (easy stage) and gradually adds high-confidence proposals later (hard stage); and (ii) a \emph{Dynamic Fusion} that progressively strengthens the contribution of historical proposals stored in a memory bank $\mathcal{B}$, improving stability and reducing false negatives. Proposals that survive in dynamic pruning are fused with historical proposals and used to supervise the multi-agent perception.

\noindent\textbf{Dynamic Pruning.} 
PPS admits pseudo labels from easy to hard by gradually raising confidence threshold. At iteration $t$, a dynamic confidence threshold 
\begin{equation}
\tau_t
= \tau_{\min}
  + (\tau_{\max} - \tau_{\min})\,
    \sigma\!\big(k_{\tau}(t - \beta_{\tau})\big)
\label{eq:pps_tau}
\end{equation}
is applied to proposals in $\widetilde{\mathcal{P}}_{\mathrm{m}}$, where $\sigma(\cdot)$ denotes the sigmoid function, $\tau_{\min}$ and $\tau_{\max}$ define lower and upper bounds, $k_{\tau}$ controls the slope, and $\beta_{\tau}$ specifies the transition center.  
Only proposals satisfying $c_i \ge \tau_t$ are retained, enabling early recall (small $\tau_t$) and later precision (large $\tau_t$).

\noindent \textbf{Dynamic Fusion.}
To improve stability, PPS integrates historical support from $\mathcal{B}$ by blending historical and current confidences.
Specifically, a dynamic weight is defined as $\lambda_t=\sigma(k_{\lambda}(t-\beta_{\lambda}))$, where $k_{\lambda}$ and $\beta_{\lambda}$ controls the slope and center of the transition. Historical confidences are re-weighted as $\tilde{c}_j^{\mathrm{h}}=\lambda_t c_j^{\mathrm{h}}$ for $(b_j^{\mathrm{h}},c_j^{\mathrm{h}})\in\mathcal{B}$,
while current predictions are weighted as $\tilde{c}_i=(1-\lambda_t)c_i$ for $(b_i,c_i)\in\widetilde{\mathcal{P}}_{\mathrm{m}}$.
The final predictions are obtained via Non-Maximum Suppression (NMS):
\begin{equation}
\widehat{\mathcal{P}}_{\mathrm{m}}
= \mathrm{NMS}\!\left(
\{b_j, \tilde{c}_{j}^{\mathrm{h}}\}_{\mathcal{B}}
\cup
\{b_i, \tilde{c}_{i}\}_{\widetilde{\mathcal{P}}_{\mathrm{m}}},
\eta
\right),
\label{eq:pps_nms}
\end{equation}
where $\eta \in (0,1)$ is the rotated-IoU threshold, $(b_i, c_{i})$ denotes the current pruned proposal and $(b_j^{\mathrm{h}}, c_{j}^{\mathrm{h}})$ denotes the historical proposal in $\mathcal{B}$, $\tilde{c}_{i}$ and $\tilde{c}_{j}^{\mathrm{h}}$ are their reweighted confidences. The stabilized proposal set $\widehat{\mathcal{P}}_{\mathrm{m}}$ is appended to $\mathcal{B}$ and supervises $D_{\mathrm{m}}$ at the next iteration.

\subsection{Cross-View Consensus Learning}
\label{sec:ccl}

As shown in Fig.~\ref{fig:overall}, our framework includes two branches: single-agent (top) and multi-agent (bottom) branches.  The single-agent branch may miss distant or occluded objects, while filtered multi-view proposals provide complementary geometric and contextual cues. {Cross-View Consensus Learning (CCL)} transfers these cues into $D_{\mathrm{e}}$ through {Multi-View Geometric Consensus} and {BEV Semantic Alignment}.

\noindent\textbf{Multi-View Geometric Consensus.}
At iteration $t$, PPF produces filtered proposal sets $\widetilde{\mathcal{P}}_{\mathrm{e}}$ and $\widetilde{\mathcal{P}}_{\mathrm{m}}$.  
We establish geometric consistency between branches by first identifying cross-view matches under a rotated-IoU threshold $\eta_{\mathrm{ccl}}\in(0,1)$, then incorporating unmatched but spatially valid multi-view proposals. The unmatched valid set is 
 
\begin{equation}
\begin{aligned}
\mathcal{U} =
\Big\{
(b_j^m, \cdot) \in \widetilde{\mathcal{P}}_m
\;\Big|\;
&\forall (b_i^e, \cdot) \in \widetilde{\mathcal{P}}_e,\;
\mathrm{IoU}(b_i^e, b_j^m) < \eta_{ccl}, \\
&\pi(b_j^m; X^e) \ge \rho
\Big\},
\end{aligned}
\label{eq:ccl_unmatched_refined}
\end{equation}
where $\pi(b;X^e)$ counts the number of points from $X^e$ inside box $b$, 
and $\rho$ is a minimal point-support threshold ensuring geometric plausibility.  
The final consensus pseudo labels are then obtained by 
$\widehat{\mathcal{P}}_e=\mathrm{NMS}\big(\widetilde{\mathcal{P}}_{\mathrm{e}}\cup\mathcal{U},\,\eta_{\mathrm{ccl}}\big)$,  
which fuses ego proposals with valid unmatched multi-view proposals through rotated-IoU NMS.

\noindent\textbf{BEV Semantic Alignment.}
While geometric matching captures spatial and geometric structures, it may overlook high-level contexts under partial visibility. To encourage higher-level context consistency, we further align BEV semantics between two branches.  
Let $\mathbf{F}_{\mathrm{e}}$ and $\mathbf{F}_{\mathrm{m}}\in\mathbb{R}^{H\times W\times C}$ denote BEV feature maps of the single-agent and multi-agent detectors, both expressed in the same coordinate system.   
A visibility mask $M\in\{0,1\}^{H\times W}$ is constructed from $\mathbf{F}_{\mathrm{e}}$ to exclude empty regions where the mean activation is below a threshold $\gamma$ by 
\begin{equation}
M(i,j)
=
\Phi \!\left(
\frac{1}{C}\sum_{z=1}^{C}\mathbf{F}_{\mathrm{e}}(i,j,z)
\ge
\gamma
\right),
\label{eq:ccl_mask}
\end{equation}
where $\Phi(\cdot)$ denotes the indicator function that outputs $1$ if the bracket condition holds and $0$ otherwise. The alignment loss is defined as
\begin{equation}
\mathcal{L}_{\mathrm{bev}}
=
\frac{1}{Z}\,
\big\|
\big(\mathbf{F}_{\mathrm{e}}-\mathbf{F}_{\mathrm{m}}\big)\odot M 
\big\|_{2}^{2},
\label{eq:ccl_bev}
\end{equation}
where $\odot$ denotes element-wise multiplication and $Z=\sum_{i,j}M(i,j)$ normalizes over valid BEV cells.

\begin{algorithm}[t]
\caption{Training Procedure of UMS}
\label{alg:ums}
\begin{algorithmic}[1]

\State \textbf{Input:} Multi-agent Point Clouds: $\{X^v\}_{v=1}^V$, iterations: $T$, epochs per iteration: $E$.
\State \textbf{Output:} Trained detectors $D_{\mathrm{m}}$ and $D_{\mathrm{e}}$.

\State Initialize the weak  $D_{\mathrm{m}}, D_{\mathrm{e}}$ and memory bank $\mathcal{B}$.

\For{iteration $t=1$ to $T$}

    \State Generate  $\mathcal{P}_{\mathrm{m}}, \mathcal{P}_{\mathrm{e}}$ from  $D_{\mathrm{m}}(\{X^v\})$ and $D_{\mathrm{e}}(X^{e})$.
    
   \Statex \hspace*{1.5em}{\footnotesize\color{gray}{\textit{// PPF  }}}

    \State Train a PointNet++  $C_{\phi}$ with Eq.~(\ref{eq:ppf_loss}) {if} $t=1$.
    \State Compute $q_i$ via Eq.~(\ref{eq:ppf}) and filter to obtain $\widetilde{\mathcal{P}}_{\mathrm{e}}, \widetilde{\mathcal{P}}_{\mathrm{m}}$.
      \Statex \hspace*{1.5em}{\footnotesize\color{gray}{\textit{// PPS }}}
    \State Dynamic Pruning with Eq.~(\ref{eq:pps_tau}). 
    \State Get $\widehat{\mathcal{P}}_{\mathrm{m}}$ by Dynamic Fusion (Eq.~(\ref{eq:pps_nms})) and update $\mathcal{B}$.

    \Statex \hspace*{1.5em}{\footnotesize\color{gray}{\textit{// CCL  }}}
    \State Get $\widehat{\mathcal{P}}_{\mathrm{e}}$ by Multi-View Geometric Consensus with 
    \hspace*{1.5em}Eq.(\ref{eq:ccl_unmatched_refined}) and NMS. 
    
    \State BEV Semantic Alignment with Eqs.~(\ref{eq:ccl_mask}),(\ref{eq:ccl_bev}).

     \Statex \hspace*{1.5em}{\footnotesize\color{gray}{\textit{// Optimization with Pseudo Labels}}}
    \For{epoch $e=1$ to $E$}
        \State Update $D_{\mathrm{m}}$ using  $\widehat{\mathcal{P}}_{\mathrm{m}}$, $\mathcal{L}_{\mathrm{m}}$ with Eq.~(\ref{eq:loss_m}).
        \State Update $D_{\mathrm{e}}$ using $\widehat{\mathcal{P}}_{\mathrm{e}}$,  $\mathcal{L}_{\mathrm{e}}$ with Eq.~(\ref{eq:loss_e}).
    \EndFor

\EndFor

\State \Return $D_{\mathrm{m}}$, $D_{\mathrm{e}}$ 

\end{algorithmic}
\end{algorithm}

\begin{table*}[t!]
\centering
\renewcommand{\arraystretch}{1.1}

\setlength{\tabcolsep}{5pt}
\caption{
\textbf{Unsupervised Multi-agent and Single-agent Perception (3D Object Detection).}
AP@0.3 / AP@0.5 on the V2V4Real~\cite{xu2023v2v4real} and OPV2V~\cite{xu2022opv2v} test sets.
}
\vspace{-0.5em}

\begin{tabular}{l|cccc|cccc}
\toprule
\multirow{3}{*}{Method} &
\multicolumn{4}{c|}{V2V4Real~\cite{xu2023v2v4real}} &
\multicolumn{4}{c}{OPV2V~\cite{xu2022opv2v}} \\

& \multicolumn{2}{c}{Multi-Agent} & \multicolumn{2}{c|}{Single-Agent} 
& \multicolumn{2}{c}{Multi-Agent} & \multicolumn{2}{c}{Single-Agent} \\
\cmidrule(lr){2-3}\cmidrule(lr){4-5}
\cmidrule(lr){6-7}\cmidrule(lr){8-9}
& AP@0.3 & AP@0.5 & AP@0.3 & AP@0.5
& AP@0.3 & AP@0.5 & AP@0.3 & AP@0.5 \\
\midrule
Supervised
& 71.35 & 64.75 & 57.40 & 50.17
& 94.80 & 94.11 & 80.01 & 77.89 \\
\hline
DBSCAN~\cite{ester1996density}
& 14.59 & 7.57 & 11.63 & 6.55
& 32.31 & 24.43 & 26.49 & 21.72\\

OYSTER~\cite{zhang2023towards}
& 37.50 & 23.52 & 29.08 & 24.25
& 56.58 & 49.01 & 42.62 & 41.93 \\

CPD~\cite{wu2024commonsense}
& 40.67 & 30.27 & 37.41 & 30.28
& 59.17 & 50.49 & 44.27 & 43.25 \\

DOtA~\cite{xia2025learning}
& 54.60 & 48.84 & 45.40 & 40.41
& 66.14 & 52.37 & 59.01 & 46.87 \\

\textbf{UMS (Ours)}
& \textbf{58.12} & \textbf{52.03} & \textbf{49.72} & \textbf{44.27}
& \textbf{86.71} & \textbf{83.89} & \textbf{76.31} & \textbf{71.30} \\
\bottomrule
\end{tabular}

\label{tab:main_results_merged}
\vspace{-1em}
\end{table*}

\subsection{Training Procedure}
The framework is trained for $T$ refinement iterations, where each iteration contains $E$ optimization epochs. In iteration $t$, the proposed modules generate two refined pseudo-label sets: (i) the ego-view consensus labels $\widehat{\mathcal{P}}_{\mathrm{e}}$ from CCL and (ii) the stabilized multi-view labels $\widehat{\mathcal{P}}_{\mathrm{m}}$ from PPS, which supervise $D_{\mathrm{e}}$ and $D_{\mathrm{m}}$ respectively. We adopt the focal loss for classification $\mathcal{L}_{\mathrm{cls}}$ and the smooth L1 loss for regression $\mathcal{L}_{\mathrm{reg}}$ following prior works~\cite{xu2022opv2v,hu2022where2comm}, weighted by coefficients $\mu_1$ and $\mu_2$ (both set to $1$ by default). The single-agent branch additionally incorporates the BEV alignment loss $\mathcal{L}_{\mathrm{bev}}$ from Eq.~(\ref{eq:ccl_bev}), weighted by $\mu_3$:

\begin{equation}
\mathcal{L}_{\mathrm{m}}
=
\mu_1\,\mathcal{L}_{\mathrm{cls}}(\widehat{\mathcal{P}}_{\mathrm{m}})
+
\mu_2\,\mathcal{L}_{\mathrm{reg}}(\widehat{\mathcal{P}}_{\mathrm{m}}), 
\label{eq:loss_m}
\end{equation}
\begin{equation}
\mathcal{L}_{\mathrm{e}}
=
\mu_1\,\mathcal{L}_{\mathrm{cls}}(\widehat{\mathcal{P}}_{\mathrm{e}})
+
\mu_2\,\mathcal{L}_{\mathrm{reg}}(\widehat{\mathcal{P}}_{\mathrm{e}})
+
\mu_3\,\mathcal{L}_{\mathrm{bev}},
\label{eq:loss_e}
\end{equation}
where $\mathcal{L}_{*}(\cdot)$ is to compute the loss function $*$ with the corresponding pseudo labels. The two detectors are optimized separately using $\mathcal{L}_{\mathrm{m}}$ and $\mathcal{L}_{\mathrm{e}}$  within each iteration. Algorithm~\ref{alg:ums} summarizes the complete training procedure.

\section{Experiment}
In this section, we evaluate the proposed UMS framework on both multi-agent and single-agent 3D perception under unsupervised setting. We first introduce the datasets, evaluation protocols, and baselines, followed by main results and detailed ablation studies.

\subsection{Experimental Setups}
\noindent\textbf{Datasets.}
We evaluate UMS on two collaborative perception benchmarks, V2V4Real~\cite{xu2023v2v4real} and OPV2V~\cite{xu2022opv2v}. 
V2V4Real is a real-world dataset collected by two LiDAR-equipped vehicles in urban and highway environments, containing around 20k LiDAR frames and 240k annotated 3D boxes. 
OPV2V is a synthetic benchmark built on CARLA and OpenCDA with 11{,}464 LiDAR frames across more than 70 scenes. 
Following the same setting  of~\cite{xu2023v2v4real,xu2022opv2v}, we focus on the vehicle class, which is consistently supported in both datasets and provides reliable annotations for evaluation.
All experiments follow the official train/test splits.

\noindent\textbf{Evaluation Metrics.}
We follow the official protocols of V2V4Real~\cite{xu2023v2v4real} and OPV2V~\cite{xu2022opv2v}, reporting 3D Average Precision (AP) at IoU thresholds of 0.3 and 0.5. 
All results are computed within the perception ranges defined by each benchmark. 
In addition to detection performance on the test set, we further evaluate pseudo-label quality on the training set using precision and recall at a fixed IoU threshold, as well as AP at IoU 0.3 and 0.5.

\noindent\textbf{Implementation Details.}
We compare UMS with representative unsupervised baselines used in collaborative perception, including DBSCAN~\cite{ester1996density}, OYSTER~\cite{zhang2023towards}, CPD~\cite{wu2024commonsense}, and DOtA~\cite{xia2025learning}. 
All methods adopt the same pose-based initialization, where candidate proposals are generated from communicated vehicle poses without manual annotations. 
PointPillars~\cite{lang2019pointpillars} is used as the detector backbone, and AttFuse~\cite{xu2022opv2v} is used for cooperative feature fusion. 
UMS refines multi-view proposals through PPF and PPS, and improves the single-agent detector using CCL. Candidate proposals are generated with a minimum confidence threshold of $0.01$.
UMS is trained with $T=20$ refinement iterations and $E=10$ epochs per iteration. We evaluate UMS under two settings in testing: 

\begin{itemize}
    \item \textit{Multi-agent Setting}: 
    ego-agent point clouds with communicated agents' shared point clouds as the model input.
    
    \item \textit{Single-agent Setting}: ego-agent point clouds only as the model input.  
\end{itemize}

\begin{figure*}[t!]
    \centering
    \includegraphics[width=\textwidth]{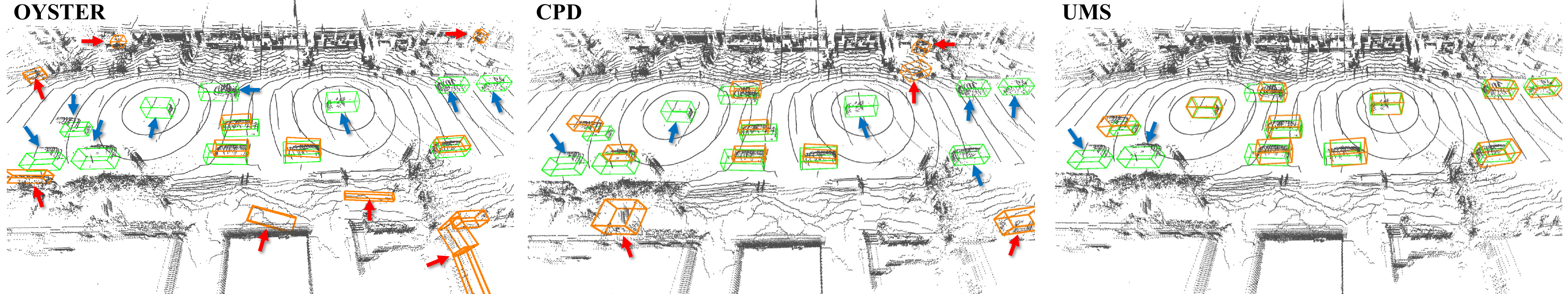}
    \caption{
\textbf{Qualitative Comparison of Pseudo Labels on V2V4Real Training Set.} Green boxes: ground truths, Orange boxes: pseudo labels, Red arrows: false positives, Blue arrows: false negatives. Multi-agent dense fused point clouds are shown here.}
    \label{fig:qualitative}
\end{figure*}

\subsection{Main Results}
\noindent\textbf{Multi-Agent Perception.}
Table~\ref{tab:main_results_merged} summarizes cooperative 3D detection results on V2V4Real and OPV2V. 
UMS achieves the best performance among all unsupervised methods. 
On OPV2V, UMS reaches 83.89\% AP@0.5, surpassing DOtA by 31.52 points, while on V2V4Real it improves AP@0.5 to 52.03\% with a gain of 3.19 points. 
The much larger improvement on OPV2V primarily comes from the clean and noise-free nature of this synthetic dataset: the hierarchical point cloud features learned by PPF can more reliably capture instance-level structure and distinguish vehicles from background regions. 
In contrast, V2V4Real contains noisy, sparse, and irregular real-world LiDAR returns, which makes instance-level feature learning more challenging and therefore limits the improvement margin. 
Clustering-based methods (DBSCAN~\cite{ester1996density}, OYSTER~\cite{zhang2023towards}, CPD~\cite{wu2024commonsense}) generally fall behind on both datasets because they lack instance-level reasoning and struggle with partial visibility and background clutter. The multi-agent comparison results in Table~\ref{tab:main_results_merged} are reported from the DOtA paper~\cite{xia2025learning}.

\noindent\textbf{Single-Agent Perception.}
UMS also improves single-agent perception. 
On V2V4Real, AP@0.5 increases from 40.41\% (DOtA) to 44.27\% (+3.86), while on OPV2V the improvement is much larger, from 46.87\% to 71.30\% (+24.43). 
This pattern mirrors the multi-agent setting: OPV2V’s consistent geometry and absence of sensor noise allow CCL to obtain more reliable cross-view cues, resulting in stronger guidance for single-agent detection. The range-wise results in Table~\ref{tab:v2v4real_range} show that improvements on V2V4Real primarily appear at mid and long ranges where single-view LiDAR becomes sparse. Overall, the effectiveness of hierarchical point cloud features and cross-view consensus is more pronounced in synthetic environments, yet still provides meaningful gains in real-world conditions.

\begin{table}[t]
\centering
\renewcommand{\arraystretch}{1.1}
\setlength{\tabcolsep}{5pt}
\caption{
\textbf{Unsupervised Single-agent Perception (3D Object Detection) in Different Ranges.}
AP@0.3 / AP@0.5 results in ranges on V2V4Real~\cite{xu2023v2v4real}. short: 0-30m, mid: 30-50m, long: 50-100m.}
\vspace{-0.5em}
\begin{tabular}{l|ccc}
\toprule
\multirow{2}{*}{Method} 
& \multicolumn{3}{c}{AP@0.3 / AP@0.5} \\
& 0--30m & 30--50m & 50--100m \\
\midrule
Supervised 
& 80.19/75.42 & 41.56/34.92 & 10.77/8.93 \\

\hline
DBSCAN~\cite{ester1996density} 
& 24.23/14.67 & 2.13/0.73 & 0.00/0.00 \\

OYSTER~\cite{zhang2023towards} 
& 50.57/42.59 & 3.82/3.02 & 0.00/0.00 \\

CPD~\cite{wu2024commonsense} 
& 62.35/50.18 & 11.57/10.06 & 0.00/0.00 \\

DOtA~\cite{xia2025learning} 
& 70.15/60.73 & 34.77/25.00 & 7.77/5.23 \\

\textbf{UMS (Ours)} 
& \textbf{70.26/65.66} & \textbf{36.26/30.05} & \textbf{9.03/7.74} \\
\bottomrule
\end{tabular}
\label{tab:v2v4real_range}
\end{table}

\begin{table}[ht]
\centering
\renewcommand{\arraystretch}{1.1}

\setlength{\tabcolsep}{5pt}
\caption{
\textbf{Pseudo-label Quality.}
Recall / Precision at IoU=0.5 on V2V4Real~\cite{xu2023v2v4real} and OPV2V~\cite{xu2022opv2v} under multi-agent setting.}
\vspace{-0.5em}
\begin{tabular}{l|cc|cc}
\toprule
\multirow{2}{*}{Method} &
\multicolumn{2}{c|}{V2V4Real~\cite{xu2023v2v4real}} &
\multicolumn{2}{c}{OPV2V~\cite{xu2022opv2v}} \\
& Recall & Precision & Recall & Precision \\
\midrule
DBSCAN~\cite{ester1996density}
& 19.59 & 2.51 
& 39.54 & 11.83 \\


OYSTER~\cite{zhang2023towards}
& 29.32 & 15.74
& 43.72 & 54.89 \\

CPD~\cite{wu2024commonsense}
& 32.15 & 16.46
& 45.80 & 56.23 \\
DOtA~\cite{xia2025learning}
& 43.91 & 60.42
& 51.87 & 65.74 \\

\textbf{UMS (Ours)}
& \textbf{53.71} & \textbf{85.98}
& \textbf{70.21} & \textbf{90.25} \\
\bottomrule
\end{tabular}
\label{tab:pseudo_quality}
\vspace{-1.5em}
\end{table}

\begin{table*}[t]
\centering
\renewcommand{\arraystretch}{1.1}
\setlength{\tabcolsep}{5pt}

\caption{
\textbf{Ablation Study of Each Component.}
Unsupervised 3D object detection on V2V4Real~\cite{xu2023v2v4real} and OPV2V~\cite{xu2022opv2v}.
}
\vspace{-0.5em}
\begin{tabular}{ccc|c c c c|c c c c}
\toprule
\multirow{2}{*}{PPF} &
\multirow{2}{*}{PPS} &
\multirow{2}{*}{CCL} &
\multicolumn{4}{c|}{V2V4Real~\cite{xu2023v2v4real}} &
\multicolumn{4}{c}{OPV2V~\cite{xu2022opv2v}} \\
\cmidrule(lr){4-7} \cmidrule(lr){8-11}
& & &
\multicolumn{2}{c}{Multi-Agent} &
\multicolumn{2}{c|}{Single-Agent} &
\multicolumn{2}{c}{Multi-Agent} &
\multicolumn{2}{c}{Single-Agent} \\
\cmidrule(lr){4-5} \cmidrule(lr){6-7} \cmidrule(lr){8-9} \cmidrule(lr){10-11}
& & &
AP@0.3 & AP@0.5 &
AP@0.3 & AP@0.5 &
AP@0.3 & AP@0.5 &
AP@0.3 & AP@0.5 \\
\midrule
 & & &
17.86 & 16.87 &
12.43 & 10.66 &
28.98 & 19.33 &
16.47 & 14.62 \\
\checkmark & & &
52.35 & 46.02 &
39.24 & 36.78 &
70.62 & 59.55 &
63.12 & 45.98 \\
\checkmark & \checkmark & &
\textbf{58.12} & \textbf{52.03} &
45.10 & 41.20 &
\textbf{86.71} & \textbf{83.89} &
74.85 & 66.44 \\
\checkmark & \checkmark & \checkmark &
{-} & {-} &
\textbf{49.72} & \textbf{44.27} &
- & - &
\textbf{76.31} & \textbf{71.30} \\
\bottomrule
\end{tabular}

\label{tab:abl_all}
\vspace{-0.8em}
\end{table*}

\noindent\textbf{Pseudo Label Quality.}
We evaluate pseudo labels on the training sets of OPV2V and V2V4Real using precision and recall at IoU\,=\,0.5 (Table~\ref{tab:pseudo_quality}). 
UMS produces pseudo labels with a clear and significant advantage over all existing unsupervised methods. UMS exhibits slightly larger precision improvements compared with recall, demonstrating strong and balanced enhancements in pseudo-label quality.
This is primarily due to PPF, which leverages cooperative point cloud density and hierarchical point-level cues to remove proposals lacking sufficient geometric support. 
The effect is more pronounced on OPV2V, where noise-free simulated LiDAR provides cleaner point patterns, making vehicle–background separation easier. 
On V2V4Real, improvements persist but are moderated by real-world sparsity, occlusion, and sensor artifacts that inherently complicate proposal classification.

\noindent\textbf{Qualitative Analysis.}
Figure~\ref{fig:qualitative} compares pseudo labels produced by OYSTER, CPD, and UMS on the V2V4Real training set. 
As indicated in the figure, OYSTER generates many false positives (red arrows), largely due to its clustering-based grouping of structured background points into vehicle-like shapes. 
CPD reduces some of these false positives but still misses a substantial number of true vehicles (blue arrows) because partial visibility and sparse LiDAR returns lead to fragmented proposals. 
In contrast, UMS produces significantly cleaner pseudo labels: false positives are almost entirely removed, and only two missed objects remain in the shown examples. 
This aligns with the quantitative precision–recall statistics in Table~\ref{tab:pseudo_quality}, demonstrating that UMS provides both higher precision and more complete object coverage than prior unsupervised methods.

\begin{figure}[t!]
    \centering
    \includegraphics[width=0.475\textwidth]{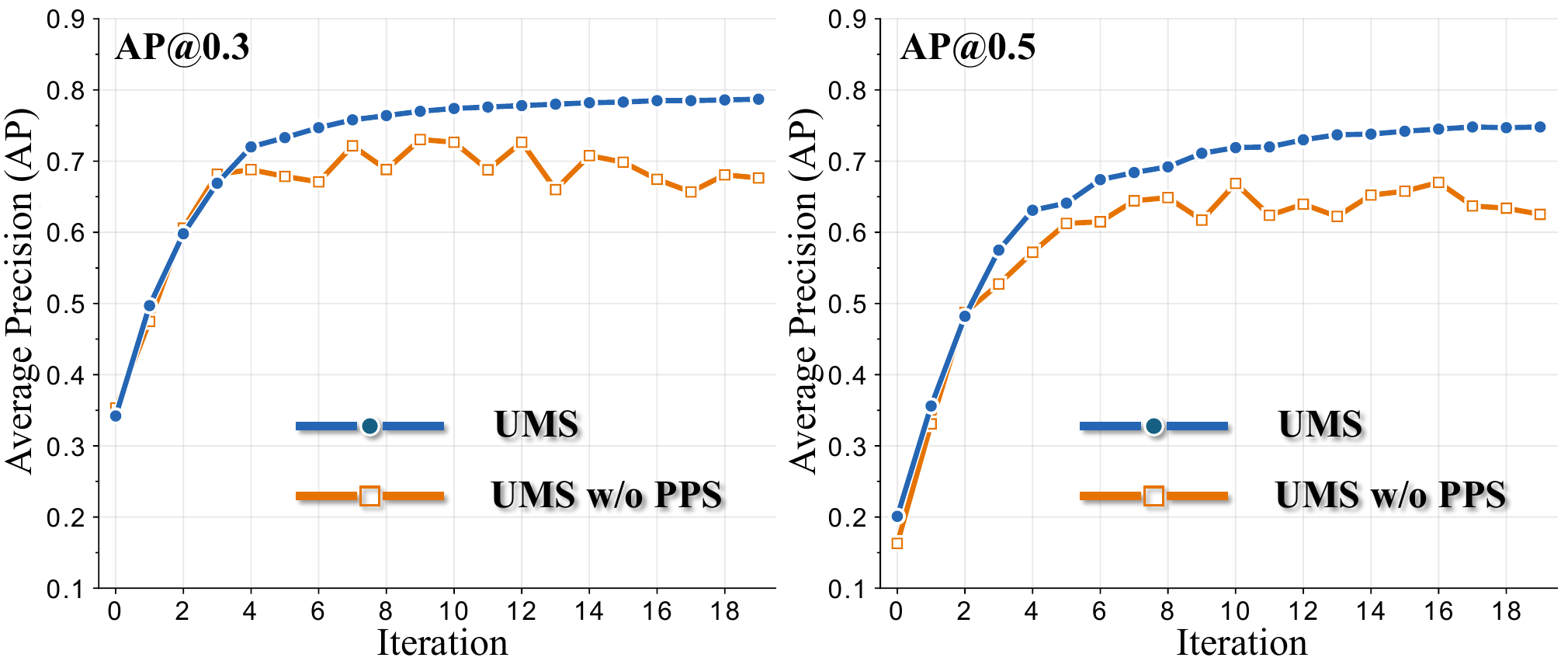}
    \caption{
    \textbf{Effect of PPS on Pseudo-Label Quality.}
Pseudo-label AP@0.3 / AP@0.5 curves across refinement iterations on OPV2V~\cite{xu2022opv2v} under the multi-agent setting.}
    \label{fig:iter_curve}
    \vspace{-1em}
\end{figure}

\subsection{Ablation Study and Discussion}
\noindent\textbf{Ablation Study of Each Component.} 
Table~\ref{tab:abl_all} summarizes the incremental contribution of the three modules. Starting from the weak detector initialized only from the prior knowledge of communicated agents, performance is notably limited due to many false positives and missed objects. Adding PPF brings the largest single improvement, increasing AP@0.5 by roughly 30-40\% on OPV2V and over 25\% on V2V4Real by suppressing proposals without sufficient geometric support. Introducing PPS further improves AP@0.5 by a noticeable margin, mainly by stabilizing intermittently visible objects across iterations. Finally, CCL adds another 3–5\% in the single-agent setting by enforcing geometric and semantic consistency between cooperative and single views. Overall, PPF and PPS drive the majority of improvements for multi-agent perception, while CCL provides the final gains for single-agent perception.

\noindent\textbf{Effect of PPS.}
Figure~\ref{fig:iter_curve} shows that PPS yields smoother and more stable improvements in pseudo-label quality on the OPV2V training set, whereas removing PPS leads to slower gains and larger fluctuations.
Table~\ref{tab:pps_tau_lambda} further evaluates the resulting multi-agent detection performance and shows that adaptive $\tau$ achieves a better balance between retaining valid proposals and suppressing noise than fixed thresholds.

\begin{table}[t]
\centering
\caption{
\textbf{Effect of $\tau$ in PPS.}
Evaluated on OPV2V~\cite{xu2022opv2v} under the multi-agent setting.
}
\vspace{-0.5em}
\begin{tabular}{l|c|cc}
\toprule
Setting & $\tau$ & AP@0.3 & AP@0.5 \\
\midrule
Low $\tau$      & 0.01   & 77.45 & 71.98 \\
High $\tau$     & 0.20   & 82.21 & 79.51 \\
Dynamic $\tau$ (ours) & sigmoid & \textbf{86.71} & \textbf{83.89} \\
\bottomrule
\end{tabular}
\label{tab:pps_tau_lambda}
\end{table}

\noindent\textbf{Effect of CCL.}
Table~\ref{tab:tau3} shows the influence of the BEV alignment weight $\mu_{3}$ in CCL. 
A small $\mu_{3}$ results in insufficient alignment, while a large value over-constrains the BEV features and introduces noise. 
The best performance is obtained at $\mu_{3}=1.5$, suggesting that a moderate alignment strength is most effective. 

\begin{table}[t]
\centering

\caption{
\textbf{Effect of $\mu_{3}$ in CCL.}
Evaluated on OPV2V~\cite{xu2022opv2v} under the single-agent setting.
}
\vspace{-0.5em}
\begin{tabular}{c|cccc}
\toprule
$\mu_3$  & $0.5$ & $1.0$ & $1.5$ & $2.0$  \\
\midrule
AP@0.3 & 75.23 & 75.81 & \textbf{76.31} & 75.44 \\
AP@0.5 & 68.65 & 70.36 & \textbf{71.30} & 71.17 \\
\bottomrule
\end{tabular}
\label{tab:tau3}
\end{table}

\noindent\textbf{Effect of Iterations.}
Table~\ref{tab:iter_rounds} reports how multi-agent detection performance changes with the number of refinement iterations. 
Both AP@0.3 and AP@0.5 increase steadily in the early stages as more partially visible objects are detected, while later iterations mainly help stabilize proposal consistency. 
Performance saturates around 15--20 iterations, indicating convergence and showing that additional refinement yields only marginal gains.

\begin{table}[t]
\centering
\caption{
\textbf{Effect of Iteration $T$.}
Evaluated on OPV2V~\cite{xu2022opv2v} under the multi-agent setting.}
\vspace{-0.5em}

\begin{tabular}{l|cccccc}
\toprule
Iteration $T$ & 1 & 5 & 10 & 20  & 25 \\
\midrule
AP@0.3 & 43.07 & 77.76 & 83.50 & {86.71} & \textbf{86.91} \\
AP@0.5 & 36.68 & 71.43 & 79.32 & \textbf{83.89} & {83.80} \\
\bottomrule
\end{tabular}
\label{tab:iter_rounds}
\vspace{-0.5em}
\end{table}

\noindent \textbf{Extension to Multi-Class Detection.}
We also conduct additional experiments on the V2X-Real dataset~\cite{xiang2024v2x} for Car and Pedestrian classes. We pretrain the filter of PPF with the Waymo open dataset~\cite{mei2022waymo} that includes labeled Car and Pedestrian point clouds and use CPD~\cite{wu2024commonsense} to generate candidate Car and Pedestrian proposals. As shown in Table~\ref{tab:diverse_classes}, UMS consistently outperforms other unsupervised methods for both classes.

\begin{table}[t]
\centering
\caption{\textbf{Multi-Class Detection}. AP@0.3 on V2X-Real~\cite{xiang2024v2x} under the multi-agent setting.}
\vspace{-0.5em}
\label{tab:diverse_classes}
\begin{tabular}{lccc}
\toprule
{{Class}} & {{CPD~\cite{wu2024commonsense}}} & {{DOtA~\cite{xia2025learning}}} & {{{UMS (Ours)}}} \\
\midrule
{Car} & {27.56} & {34.27} & {\textbf{40.10}} \\
{Pedestrian} & {11.37} & {14.33} & {\textbf{17.71}} \\
\bottomrule
\end{tabular}
\vspace{-0.8em}
\end{table}

\noindent \textbf{Robustness under Localization Errors and Latency.}
We evaluate robustness on V2V4Real by adding Gaussian noise (std = 0.2\,m) to GPS poses and simulating a 100\,ms communication delay. 
As shown in Table~\ref{tab:v2v4real_robustness}, UMS achieves the best performance across all scenarios.  The comparison results in Table~\ref{tab:v2v4real_robustness} are obtained by retraining their models.

\begin{table}[t]
\centering

\caption{\textbf{Robustness.} AP@0.3/0.5 on V2V4Real~\cite{xu2023v2v4real} with GPS pose errors,  communication latency under the multi-agent setting.}
\vspace{-0.5em}
\label{tab:v2v4real_robustness}
\resizebox{\columnwidth}{!}{%
\begin{tabular}{lccc}
\toprule
\textbf{Method} & \textbf{Normal} & \textbf{GPS Pose Error} & \textbf{Latency} \\
\midrule
Supervised & 71.35 / 64.75 & 68.71 / 57.70 & 68.72 / 57.72 \\
\midrule 
DBSCAN~\cite{ester1996density}     & 14.42 / 7.01  & 13.63 / 6.78  & 14.45 / 7.05  \\
OYSTER~\cite{zhang2023towards}     & 36.64 / 25.87 & 35.81 / 24.81 & 36.39 / 25.48 \\
CPD~\cite{wu2024commonsense}        & 39.38 / 31.88 & 38.84 / 29.69 & 39.23 / 30.51 \\
DOtA~\cite{xia2025learning}       & 54.60 / 48.84 & 53.23 / 45.76       & 54.20 / 45.23  \\
\textbf{UMS (Ours)} & \textbf{58.12 / 52.03} & \textbf{56.21 / 49.05} & \textbf{57.67 / 48.38} \\
\bottomrule
\end{tabular}%
}
\vspace{-0.5em}
\end{table}

\section{Conclusion}
\label{sec:conclusion}
In this paper, we investigated the challenge of unsupervised 3D object detection in the multi-agent and single-agent perception. To overcome this limitation, we introduced UMS, the first unsupervised framework that jointly tackles both multi-agent and single-agent 3D perception by leveraging cooperative LiDAR during training without any human annotations. UMS integrates three key ideas: learning hierarchical point cloud features to suppress background-induced proposal errors, stabilizing multi-view proposals across refinement iterations, and enforcing cross-view consensus (geometric and BEV consistency) to guide the ego detector. Together, these components form robust collaborative guidance that yields higher-quality pseudo labels and improved 3D detection performance. Experiments on the public V2V4Real and OPV2V demonstrate clear improvements over other existing unsupervised methods.

{
    \small
    \bibliographystyle{ieeenat_fullname}
    \bibliography{main}

@inproceedings{ester1996density,
  title={A density-based algorithm for discovering clusters in large spatial databases with noise},
  author={Ester, Martin and Kriegel, Hans-Peter and Sander, J{\"o}rg and Xu, Xiaowei and others},
  booktitle={kdd},
  volume={96},
  number={34},
  pages={226--231},
  year={1996}
}

@article{li2021learning,
  title={Learning distilled collaboration graph for multi-agent perception},
  author={Li, Yiming and Ren, Shunli and Wu, Pengxiang and Chen, Siheng and Feng, Chen and Zhang, Wenjun},
  journal={Advances in Neural Information Processing Systems},
  volume={34},
  pages={29541--29552},
  year={2021}
}

@inproceedings{you2022learning,
  title={Learning to detect mobile objects from lidar scans without labels},
  author={You, Yurong and Luo, Katie and Phoo, Cheng Perng and Chao, Wei-Lun and Sun, Wen and Hariharan, Bharath and Campbell, Mark and Weinberger, Kilian Q},
  booktitle={IEEE/CVF Conference on Computer Vision and Pattern Recognition},
  pages={1130--1140},
  year={2022}
}

@article{wang20224d,
  title={4d unsupervised object discovery},
  author={Wang, Yuqi and Chen, Yuntao and Zhang, Zhao-Xiang},
  journal={Advances in Neural Information Processing Systems},
  volume={35},
  pages={35563--35575},
  year={2022}
}

@inproceedings{xu2022opv2v,
  title={Opv2v: An open benchmark dataset and fusion pipeline for perception with vehicle-to-vehicle communication},
  author={Xu, Runsheng and Xiang, Hao and Xia, Xin and Han, Xu and Li, Jinlong and Ma, Jiaqi},
  booktitle={IEEE International Conference on Robotics and Automation},
  pages={2583--2589},
  year={2022},
  organization={IEEE}
}

@inproceedings{lei2022latency,
  title={Latency-aware collaborative perception},
  author={Lei, Zixing and Ren, Shunli and Hu, Yue and Zhang, Wenjun and Chen, Siheng},
  booktitle={European Conference on Computer Vision},
  pages={316--332},
  year={2022},
  organization={Springer}
}

@article{xiang2022v2xp,
  title={V2xp-asg: Generating adversarial scenes for vehicle-to-everything perception},
  author={Xiang, Hao and Xu, Runsheng and Xia, Xin and Zheng, Zhaoliang and Zhou, Bolei and Ma, Jiaqi},
  journal={arXiv preprint arXiv:2209.13679},
  year={2022}
}

@inproceedings{najibi2022motion,
  title={Motion inspired unsupervised perception and prediction in autonomous driving},
  author={Najibi, Mahyar and Ji, Jingwei and Zhou, Yin and Qi, Charles R and Yan, Xinchen and Ettinger, Scott and Anguelov, Dragomir},
  booktitle={European Conference on Computer Vision},
  pages={424--443},
  year={2022},
  organization={Springer}
}

@inproceedings{hu2023collaboration,
  title={Collaboration helps camera overtake lidar in 3d detection},
  author={Hu, Yue and Lu, Yifan and Xu, Runsheng and Xie, Weidi and Chen, Siheng and Wang, Yanfeng},
  booktitle={IEEE/CVF Conference on Computer Vision and Pattern Recognition},
  pages={9243--9252},
  year={2023}
}

@article{wei2023asynchrony,
  title={Asynchrony-robust collaborative perception via bird's eye view flow},
  author={Wei, Sizhe and Wei, Yuxi and Hu, Yue and Lu, Yifan and Zhong, Yiqi and Chen, Siheng and Zhang, Ya},
  journal={Advances in Neural Information Processing Systems},
  volume={36},
  pages={28462--28477},
  year={2023}
}

@inproceedings{xu2023v2v4real,
  title={V2v4real: A real-world large-scale dataset for vehicle-to-vehicle cooperative perception},
  author={Xu, Runsheng and Xia, Xin and Li, Jinlong and Li, Hanzhao and Zhang, Shuo and Tu, Zhengzhong and Meng, Zonglin and Xiang, Hao and Dong, Xiaoyu and Song, Rui and others},
  booktitle={IEEE/CVF conference on computer vision and pattern recognition},
  pages={13712--13722},
  year={2023}
}

@inproceedings{zhang2023towards,
  title={Towards unsupervised object detection from lidar point clouds},
  author={Zhang, Lunjun and Yang, Anqi Joyce and Xiong, Yuwen and Casas, Sergio and Yang, Bin and Ren, Mengye and Urtasun, Raquel},
  booktitle={IEEE/CVF Conference on Computer Vision and Pattern Recognition},
  pages={9317--9328},
  year={2023}
}

@inproceedings{wu2024commonsense,
  title={Commonsense prototype for outdoor unsupervised 3d object detection},
  author={Wu, Hai and Zhao, Shijia and Huang, Xun and Wen, Chenglu and Li, Xin and Wang, Cheng},
  booktitle={IEEE/CVF Conference on Computer Vision and Pattern Recognition},
  pages={14968--14977},
  year={2024}
}

@article{lentsch2024union,
  title={Union: Unsupervised 3d object detection using object appearance-based pseudo-classes},
  author={Lentsch, Ted and Caesar, Holger and Gavrila, Dariu},
  journal={Advances in Neural Information Processing Systems},
  volume={37},
  pages={22028--22046},
  year={2024}
}

@inproceedings{hu2024communication,
  title={Communication-efficient collaborative perception via information filling with codebook},
  author={Hu, Yue and Peng, Juntong and Liu, Sifei and Ge, Junhao and Liu, Si and Chen, Siheng},
  booktitle={IEEE/CVF Conference on Computer Vision and Pattern Recognition},
  pages={15481--15490},
  year={2024}
}

@INPROCEEDINGS{li2024s2r,
  author={Li, Jinlong and Xu, Runsheng and Liu, Xinyu and Li, Baolu and Zou, Qin and Ma, Jiaqi and Yu, Hongkai},
  booktitle={IEEE International Conference on Robotics and Automation}, 
  title={S2R-ViT for Multi-Agent Cooperative Perception: Bridging the Gap from Simulation to Reality}, 
  year={2024},
  volume={},
  number={},
  pages={16374-16380},
  doi={10.1109/ICRA57147.2024.10611572}
}

@inproceedings{li2025v2x,
  title={V2x-dgw: Domain generalization for multi-agent perception under adverse weather conditions},
  author={Li, Baolu and Li, Jinlong and Liu, Xinyu and Xu, Runsheng and Tu, Zhengzhong and Guo, Jiacheng and Zou, Qin and Li, Xiaopeng and Yu, Hongkai},
  booktitle={IEEE International Conference on Robotics and Automation},
  pages={974--980},
  year={2025},
  organization={IEEE}
}

@inproceedings{xia2025learning,
  title={Learning to Detect Objects from Multi-Agent LiDAR Scans without Manual Labels},
  author={Xia, Qiming and Lin, Wenkai and Xiang, Haoen and Huang, Xun and Chen, Siheng and Dong, Zhen and Wang, Cheng and Wen, Chenglu},
  booktitle={IEEE/CVF Conference on Computer Vision and Pattern Recognition},
  pages={1418--1428},
  year={2025}
}

@article{qi2017pointnet++,
  title={Pointnet++: Deep hierarchical feature learning on point sets in a metric space},
  author={Qi, Charles Ruizhongtai and Yi, Li and Su, Hao and Guibas, Leonidas J},
  journal={Advances in Neural Information Processing Systems},
  volume={30},
  year={2017}
}

@inproceedings{bengio2009curriculum,
  title={Curriculum learning},
  author={Bengio, Yoshua and Louradour, J{\'e}r{\^o}me and Collobert, Ronan and Weston, Jason},
  booktitle={International Conference on Machine Learning},
  pages={41--48},
  year={2009}
}

@inproceedings{lang2019pointpillars,
  title={Pointpillars: Fast encoders for object detection from point clouds},
  author={Lang, Alex H and Vora, Sourabh and Caesar, Holger and Zhou, Lubing and Yang, Jiong and Beijbom, Oscar},
  booktitle={IEEE/CVF Conference on Computer Vision and Pattern Recognition},
  pages={12697--12705},
  year={2019}
}

@inproceedings{wang2020v2vnet,
  title={V2vnet: Vehicle-to-vehicle communication for joint perception and prediction},
  author={Wang, Tsun-Hsuan and Manivasagam, Sivabalan and Liang, Ming and Yang, Bin and Zeng, Wenyuan and Urtasun, Raquel},
  booktitle={European conference on computer vision},
  pages={605--621},
  year={2020},
  organization={Springer}
}

@article{hu2022where2comm,
  title={Where2comm: Communication-efficient collaborative perception via spatial confidence maps},
  author={Hu, Yue and Fang, Shaoheng and Lei, Zixing and Zhong, Yiqi and Chen, Siheng},
  journal={Advances in Neural Information Processing Systems},
  volume={35},
  pages={4874--4886},
  year={2022}
}

@article{fruhwirth2024vision,
  title={Vision-Language Guidance for LiDAR-based Unsupervised 3D Object Detection},
  author={Fruhwirth-Reisinger, Christian and Lin, Wei and Mali{\'c}, Du{\v{s}}an and Bischof, Horst and Possegger, Horst},
  journal={arXiv preprint arXiv:2408.03790},
  year={2024}
}

@inproceedings{tian2021unsupervised,
  title={Unsupervised object detection with lidar clues},
  author={Tian, Hao and Chen, Yuntao and Dai, Jifeng and Zhang, Zhaoxiang and Zhu, Xizhou},
  booktitle={IEEE/CVF Conference on Computer Vision and Pattern Recognition},
  pages={5962--5972},
  year={2021}
}

@inproceedings{mei2022waymo,
  title={Waymo open dataset: Panoramic video panoptic segmentation},
  author={Mei, Jieru and Zhu, Alex Zihao and Yan, Xinchen and Yan, Hang and Qiao, Siyuan and Chen, Liang-Chieh and Kretzschmar, Henrik},
  booktitle={European Conference on Computer Vision},
  pages={53--72},
  year={2022},
  organization={Springer}
}

@inproceedings{xiang2024v2x,
  title={V2x-real: a large-scale dataset for vehicle-to-everything cooperative perception},
  author={Xiang, Hao and Zheng, Zhaoliang and Xia, Xin and Xu, Runsheng and Gao, Letian and Zhou, Zewei and Han, Xu and Ji, Xinkai and Li, Mingxi and Meng, Zonglin and others},
  booktitle={European Conference on Computer Vision},
  pages={455--470},
  year={2024},
  organization={Springer}
}

@article{albattah2022overview,
  title={An overview of the current challenges, trends, and protocols in the field of vehicular communication},
  author={Albattah, Waleed and Habib, Shabana and Alsharekh, Mohammed F and Islam, Muhammad and Albahli, Saleh and Dewi, Deshinta Arrova},
  journal={Electronics},
  volume={11},
  number={21},
  pages={3581},
  year={2022},
  publisher={MDPI}
}

@article{huang2025vehicle,
  title={Vehicle-to-everything cooperative perception for autonomous driving},
  author={Huang, Tao and Liu, Jianan and Zhou, Xi and Nguyen, Dinh C and Azghadi, Mostafa Rahimi and Xia, Yuxuan and Han, Qing-Long and Sun, Sumei},
  journal={Proceedings of the IEEE},
  year={2025},
  publisher={IEEE}
}
}

\end{document}